# DESIGN OF A VISUAL POSE ESTIMATION ALGORITHM FOR MOON LANDING

Atakan SÜSLÜ[1)], Betül Rana KURAN[1)], Halil Ersin SÖKEN[1)]

[1)] *Department of Aerospace Engineering, Middle East Technical University, Ankara, Turkey*
*atakan.suslu@metu.edu.tr*

In order to make a pinpoint landing on the Moon, the spacecraft's navigation system must be accurate. To achieve the desired accuracy, navigational drift caused by the inertial sensors must be corrected. One way to correct this drift is to use absolute navigation solutions. In this study, a terrain absolute navigation method to estimate the spacecraft's position and attitude is proposed. This algorithm uses the position of the craters below the spacecraft for estimation. Craters seen by the camera onboard the spacecraft are detected and identified using a crater database known beforehand. In order to focus on estimation algorithms, image processing and crater matching steps are skipped. The accuracy of the algorithm and the effect of the crater number used for estimation are inspected by performing simulations.

**Key Words:** Absolute Visual Navigation, Pose Estimation, Lunar Exploration, Craters

## 1. Introduction

Landing missions to the Moon require the spacecraft to land at the selected location accurately. In order to do so, the spacecraft must first know its position and attitude correctly. It is known that the navigation system has the most responsibility for an accurate landing[1, 2)].

In order to make a pose estimation, the spacecraft uses different sensors. The most important of these sensors is the inertial measurement unit, which can sense angular velocity and linear accelerations. The importance comes from the fact that this is an internal sensor and does not depend on anything other than itself. Other sensors used in spacecraft, such as the Sun sensor or magnetometers, cannot give meaningful readings when the Sun is not visible, or the magnetic field around the spacecraft is too low. On the other hand, due to the working principle of the inertial sensor, error accumulates with time. To have an accurate pose estimation, the other sensors must correct errors of the inertial measurement unit. In order to correct these errors, the spacecraft's pose must be estimated using an absolute estimation method.

In this paper, an absolute estimation algorithm to estimate both the position and the attitude of the spacecraft is proposed. This algorithm uses the position of the craters for estimation. Craters seen by the camera onboard the spacecraft are detected and matched first. For matching, a crater database that includes the position and size information of the craters can be used. As the absolute positions of the craters are known beforehand, this algorithm can be classified as a terrain absolute navigation method[3)]. In order to focus on the estimation algorithm rather than the image processing and crater identification, they are not considered thoroughly, but modeled using the studies available in the literature.

The algorithm proposed in this paper uses the nonlinear least squares method together with the QUEST method to estimate both position and attitude using only the relative position of the craters with respect to the spacecraft and the distance between the spacecraft and craters. Estimations done by this algorithm can be used to fix the accumulated errors caused by inertial measurement errors.

For simulation, the spacecraft and the Moon is modeled using Matlab. The accuracy of the estimation algorithm and the effect of the number of craters used for estimation is inspected using the simulation results.

## 2. Crater Detection Methods

In order to get the required information to estimate the position and attitude of the spacecraft, visible craters on the Moon can be used. Using the camera onboard, the spacecraft can detect the visible craters. The unit direction vector from the spacecraft to the visible craters and the crater distances from the spacecraft can be obtained from the camera measurements. Craters can be detected using neural networks, and they can be identified with a craters database using the methods available in the literature[4, 5)]. Distance from the spacecraft to craters can be found using stereo cameras. Also, a method to find the distance from an image is available in the literature for mono cameras[6)], which uses the actual radius of craters available in the crater database.

### 2.1. Crater Database

A crater database with plenty of crater information is required to estimate position and attitude. A crater database found in literature[7)] is used as the database to compare crater formations for identification and get reference information for estimation algorithms. The database contains over 2 million craters with name, diameter, latitude, and longitude information. 1.3 million craters have diameters larger than 1 kilometer, 83000 have diameters larger than 5 kilometers, and 6972 have diameters larger than 20 kilometers.

## 3. Modelling and Simulation

### 3.1. Coordinate Frames

#### 3.1.1. Moon Centered Inertial Frame

The Moon centered inertial (MCI) frame, is a stationary coordinate frame whose origin is at the Moon's center of mass. The z-axis is along the Moon's rotation axis. The x-axis is in the equatorial plane, pointing towards the prime meridian at specified epoch. The y-axis completes the right-handed system. Fig. 1 demonstrates the axes of the MCI frame with the superscript $i$.

#### 3.1.2. Moon Centered Moon Fixed Frame

The Moon centered Moon fixed (MCMF) frame, rotates along with the Moon, whose origin is at the Moon's center of mass. The z-axis is along the Moon's rotation axis. The x-axis is in the equatorial plane, pointing towards the prime meridian. The y-axis completes the right-handed system. At the beginning of the simulation ($t_0$), MCI and MCMF frames are coincident. Fig. 1 illustrates the axes of the MCMF frame with the superscript $m$. The term $\boldsymbol{\omega}_{im}^m$ is the Moon's rotation rate with respect to the MCI frame resolved in the MCMF frame. Therefore, as time progress ($t > t_0$), the angle difference between $X^i$ and $X^m$, and between $Y^i$ and $Y^m$ occurs due to the rotation of the MCMF frame, and it is represented with $\theta_m$.

#### 3.1.3. Body Frame

The origin of the body frame coincides with the center of mass of the body. The x-axis points towards the forward direction. The y-axis points towards the transverse direction. The z-axis direction points towards the vertical direction, completing a right-handed system.

#### 3.1.4. Selenocentric Coordinate System

The origin of the selenocentric coordinate system is at the center of the mass of the Moon, and it is a Moon fixed frame. Unlike the MCMF frame, in which rectangular coordinates are used, spherical coordinates are used in the selenocentric coordinate system. The location of the point of interest is described in latitude, $\varphi$, longitude, $\lambda$, and radius, $r$ in the selenocentric coordinate system. The lunar equatorial plane is taken as the $X^m - Y^m$ plane, which is shown in Fig. 1.

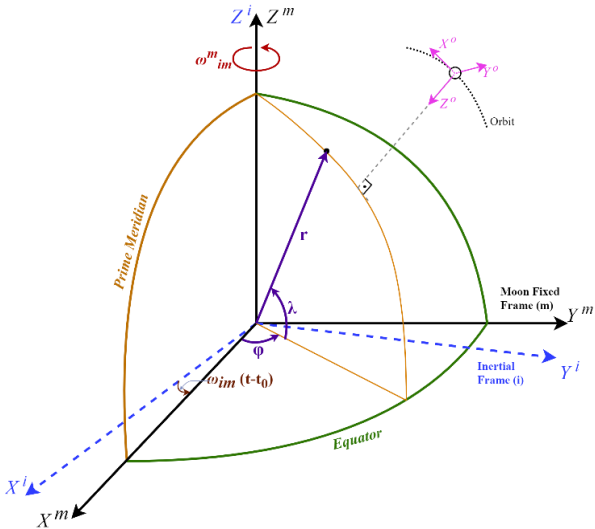

Fig. 1. Coordinate frames used for simulation.

#### 3.1.5. Orbit Frame

The origin of the orbit frame coincides with the center of mass of the body. The z-axis points towards the nadir direction, from the body to the center of the Moon. The y-axis points towards the orbit normal, and the direction is the same as the opposite direction of the body's orbital angular velocity. The x-axis direction completes a right-handed system. Orbit frame was used to define gravitational acceleration more easily. Fig. 1 demonstrates the axes of the orbit frame with the superscript $o$.

### 3.2. Modelling the Moon

While modeling, the shape of the Moon is assumed to be sphere due to the low flattening of the Moon, which is 0.0012[8]. The Moon rotates around z-axes of MCFC and MCI, since z-axis of these frames coincide, and an angle difference occurs between frames in XY-plane. This angle, which is denoted by $\theta_m$, can be determined using the equation below.

$$\theta_m = \int_{t_0}^{t} \boldsymbol{\omega}_{im}^m \, dt \quad (1)$$

DCM matrix from MCMF frame to MCI frame, $R_{im}$, can be created using $\theta_m$ as below.

$$R_{im} = \begin{bmatrix} \cos\theta_m & \sin\theta_m & 0 \\ -\sin\theta_m & \cos\theta_m & 0 \\ 0 & 0 & 1 \end{bmatrix} \quad (2)$$

This DCM is especially useful for transforming the positions of the craters from the MCMF frame to the MCI frame. Crater positions can be easily defined in the MCMF as craters also rotate with the Moon, and the positions of the craters in the MCMF are constant. However, crater positions are defined in the selenocentric frame, which uses spherical coordinates $\lambda$, $\varphi$ and $r$. In order to transfer the positions from the selenocentric frame to MCMF, equations below can be used.

$$X^m = r_M \cos\varphi \sin\lambda \quad (3a)$$
$$Y^m = r_M \cos\varphi \sin\lambda \quad (3b)$$
$$Z^m = r_M \sin\varphi \quad (3c)$$

where $r_M$ represents the radius of the Moon and is constant due to the perfect sphere assumption.

### 3.3. Modelling the Spacecraft

Spacecraft equations of motion in the MCI frame are used to calculate the actual position and attitude of the spacecraft.

$$\begin{bmatrix} \dot{\boldsymbol{r}}^i \\ \dot{\boldsymbol{v}}^i \\ \dot{\boldsymbol{q}}_{ib} \end{bmatrix} = \begin{bmatrix} \boldsymbol{v}^i \\ R_{ib}\boldsymbol{f}_b + R_{io}\boldsymbol{g}^o \\ 0.5 \cdot \boldsymbol{\Omega}(\boldsymbol{\omega}_{ib}^b)\boldsymbol{q}_{ib} \end{bmatrix} \quad (4)$$

where $\boldsymbol{r}^i$ is the position of the spacecraft with respect to the MCI frame, $\boldsymbol{v}^i$ is the velocity of the spacecraft with respect to the MCI frame and $\boldsymbol{q}_{ib}$ is the attitude quaternion of the spacecraft body frame with respect to the MCI frame. $\boldsymbol{f}_b$ is the specific force acting on the spacecraft, which is zero unless there is thruster activity. $\boldsymbol{g}^o$ is the gravitational acceleration in the orbit frame. The gravity of the Moon is modeled using Newton's law of gravitation, and gravity is in nadir direction which is the z axis of the orbit frame. Therefore, gravity acting on the spacecraft can be found as

$$\boldsymbol{g}^o = \frac{\mu}{r_b^2}\hat{\boldsymbol{k}} \quad (5)$$

where $r_b$ is the distance from the center of the Moon to the spacecraft.

Quaternions are used to simulate attitude of the spacecraft.

However, in order to transform vectors from body frame to inertial frame, DCM matrix from body frame to inertial frame, $R_{ib}$, is required. Required DCM can be obtained by converting quaternions to DCM. Detailed information about the equations of motion and quaternion-DCM transformations can be found in ref. 9 and ref. 10.

Camera is located at the bottom of the spacecraft such that the camera points at the z direction in body axis. Position and attitude of the camera in the body frame is fixed and direction that camera points can be found in inertial frame using $R_{ib}$.

### 3.4. Crater Detection

For the simulation, image processing steps are skipped for crater detection. Angle of view of the camera is taken as 45°. Four lines that are representing the camera vision limits are attached to the body frame to find out which craters are visible to the camera at each moment. Using three-dimensional analytic geometry, intersection points of the camera lines and the Moon in the inertial frame are found using equation below.

$$x = o + d\hat{u} \quad (6)$$

where $x$ is the intersection point, $o$ is the origin of the line, which is the position of the spacecraft in this case, $d$ is the distance between spacecraft and intersection point and $\hat{u}$ is the unit vector in the direction of the line. Distance between the spacecraft and the intersection point can be found as

$$d = -[\hat{u} \cdot (o - c)] \pm \sqrt{\Delta} \quad (7)$$

where $c$ is the center of the Moon, which is the origin in the MCI frame. $\Delta$ can be found as

$$\Delta = [\hat{u} \cdot (o - c)]^2 - (\|o - c\|^2 - r_M^2) \quad (8)$$

where $r_M$ is the radius of the Moon.

After finding the intersection points, coordinates of these points are converted to the selenocentric frame, which is the frame that crater positions in the database are defined. Craters between intersection points are decided to be candidate visible craters.

Visible craters from candidates are selected using the diameter information of the craters, which are available in the database. When the spacecraft is at high altitudes, craters with larger diameter can be seen and craters with smaller diameters cannot be seen. Visible craters are determined using the condition below.

$$D_{cr} > 0.1826 \cdot e^{0.01701 \cdot h} \quad (9)$$

where $D_{cr}$ is the diameter of the crater in meters and $h$ is the altitude of the spacecraft in kilometers.

Crater identification process is also modelled. Using the identification methods found in literature[5], it is decided that a visible crater has 85% chance of being identified based on the worst-case scenario of the identification method.

### 3.5. Measurement Model

It is assumed that camera gives the unit direction vector from spacecraft to each visible crater center, $\hat{r}_{bc}^b$, and distance between the spacecraft and each visible crater, $\rho_c$. To build the measurements of the camera, Gaussian white noise is added to the spacecraft model outputs of $\hat{r}_{bc,\text{actual}}^b$ and $\rho_{c,\text{actual}}$ such that the noise with the standard deviation $\sigma_d = 10^{-4}$ for unit direction vector measurements and $\sigma_\rho = 10\ m$ for range measurements. The following equations are used to simulate the camera measurement.

For the unit vector from spacecraft to each visible crater measurements:

$$\hat{r}_{bc}^b = \hat{r}_{bc,\text{actual}}^b + v_d \quad (10)$$

where $\hat{r}_{bc,\text{actual}}^b = R_{bi} \frac{r_c^i - r_b^i}{|r_c^i - r_b^i|}$ and $v_d$ is the Gaussian white noise with standard deviation of $\sigma_d$.

For the distance between the spacecraft and each visible crater measurement:

$$\rho_c = \rho_{c,\text{actual}} + v_\rho \quad (11)$$

where $\rho_{c,\text{actual}} = \|r_c^i - r_b^i\|$ and $v_\rho$ is the Guassian white noise with standard deviation of $\sigma_\rho$.

### 4. Estimation Methods

Fig. 2 shows the block diagram of the position and attitude estimation method. It was mentioned that the measurements of the camera are the range to each crater, $\rho_c$, and the unit direction vector from the spacecraft to each crater in the body frame, $\hat{r}_{bc}^b$. The sensor provides these measurements for the detected craters. Also, the position vector of the matched craters from the crater map with the detected craters in MCI frame, $r_c^i$ is determined. $\rho_c$ and $r_c^i$ are the inputs of the position estimation. For the position estimation nonlinear least square method is used. Then, by using the unit vector of the position estimation, $\hat{r}_{ib}^b$, and the unit vector of the positions of the matched craters, $\hat{r}_c^i$, the reference information for the attitude estimation, $\hat{r}_{bc}^i$, is determined. The other input of the attitude estimation, which is the unit vector from the spacecraft to each crater in the body frame, $\hat{r}_{bc}^b$, comes from the sensor. Then, attitude can be estimated by using the QUEST method.

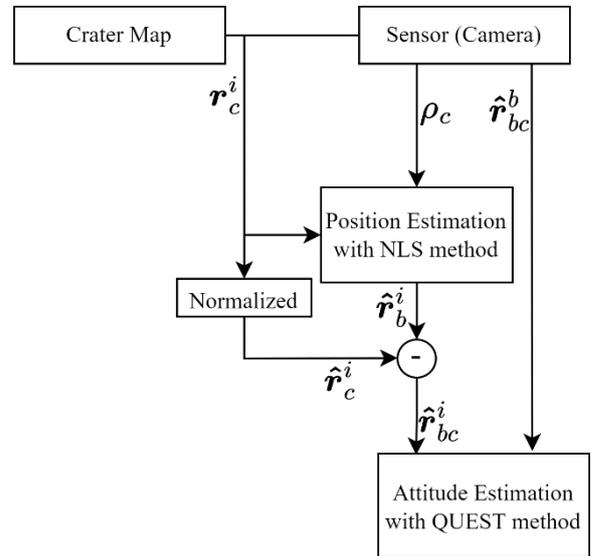

Fig. 2. Block diagram of the position and attitude estimation.

### 4.1. Position Estimation

It was mentioned that for the position estimation nonlinear least square method is used. The range data, $\rho_c$ and position data, $\hat{r}_c^i$, of the $n$ number of craters are inserted into the nonlinear least-square algorithm. Following $f(x)$ function used in NLS algorithm, derived from the range measurement. The sum of the first three term in the right side of the equation is equal to the square of the range measurement.

$$f_j(x) = (x_j - x_u)^2 + (y_j - y_u)^2 + (z_j - z_u)^2 - (\rho_c)_j \quad (12)$$

where $(r_c^i)_j = [x_j, y_j, z_j]$ represents the position vector of the $j^{th}$ crater in MCI frame where j = 1, 2, ..., n and n should be higher than 2 or else position estimation skips for that time. $(\rho_c)_j$ is the range to the $j^{th}$ crater. $r_b^i = x = [x_u, y_u, z_u]$ is the position of the spacecraft in MCI frame, which will be estimated.

Then, by using $f(x)$ function, $F(x)$ matrix is created. $J(x)$ matrix is the Jacobian of the F matrix. By using $F(x)$ and $J(x)$ matrices, the position estimation of the spacecraft in MCI frame is obtained with Newton Raphson method, which can be seen in following equations.

$$x_{k+1} = x_k - J^{-1}(x_k)F(x_k). \quad (13)$$

$$\text{where } F(x) = \begin{bmatrix} f_1(x) \\ \vdots \\ f_n(x) \end{bmatrix}, \text{ and } J(x) = \begin{bmatrix} \frac{\partial f_1}{\partial x_u} & \frac{\partial f_1}{\partial y_u} & \frac{\partial f_1}{\partial z_u} \\ \vdots & \vdots & \vdots \\ \frac{\partial f_n}{\partial x_u} & \frac{\partial f_n}{\partial y_u} & \frac{\partial f_n}{\partial z_u} \end{bmatrix} \quad (14)$$

To determine the inverse of $J(x)$ matrix $3 \times n$, pseudo-inverse is used. Moreover, the initial estimation is an important factor for successful result of Newton Raphson method. When the proper initial estimation was not made, the algorithm may fail. Therefore, to solve this problem, the previous successful position estimation was used as an initial estimation.

For $|x_{k+1} - x_k| \leq \epsilon$ where $\epsilon = 10^{-9}$, the position vector, $r_b^i$, is estimated and equal to $x_{i+1}$.

### 4.2. Attitude Estimation

Using the crater direction vectors in the body frame obtained from the camera, $r_{bc}^b$, and reference vectors in MCI frame, $r_{bc}^i$, attitude of the spacecraft is estimated using Shuster's QUEST algorithm[11]. In order to obtain the reference vector in MCI frame, the position vector of craters in the MCI frame, $r_c^i$, which is known from crater map as discussed in Section 3.3, and the position vector of the spacecraft in the MCI frame, $r_b^i$, which is known from position estimation are used as below.

$$r_{bc}^i = r_c^i - r_b^i$$

The attitude quaternion of the body frame with respect to the MCI frame can be estimated using the method described above.

### 5. Results

In order to test the estimation algorithm, a simulation of the spacecraft is conducted. In the simulation, the spacecraft is in an elliptic orbit with an apoapsis of 300 km, and the periapsis of the orbit is inside the Moon. Therefore, the spacecraft slams directly into the Moon. The orbit has an inclination of 15 degrees, and simulation starts when the spacecraft is at the periapsis. The camera points in the nadir direction to see the craters on the surface.

For visualization, 3D animation of the simulation is prepared using Matlab. An example frame from the result animations can be seen in Fig. 3. In this figure, red curved line indicates the spacecraft's path; orange, blue and yellow triad indicates the spacecraft body axes, purple dashed lines indicate camera limits, and grey circles indicate the visible craters. An important advantage of the animation is that the distribution of the visible craters on the Moon can be visualized easily. Using the animation helped authors to correctly implement the crater detection model described in Section 3.4.

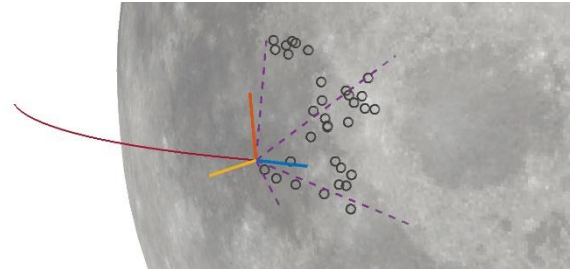

Fig. 3.   3D animation of the simulation.

As the algorithm uses craters to estimate the position and the attitude, the number of detected craters carries importance. Therefore, the number of craters visible to the camera is plotted with the altitude of the spacecraft in Fig. 4 to investigate the change in the visible crater number with respect to the altitude. It is important to notice that all the craters visible to the camera are considered as detected, and the success rate of the crater identification algorithm is not considered for this case. Therefore, the plot shows the maximum number of visible craters at each moment.

It can be seen that in the first 1500 seconds, the detected crater number is lower than 150. Between the 1500th and 2850th seconds, the detected crater number increases significantly and peaks with approximately 900 craters. After 2850 seconds, the detected crater number decreases sharply.

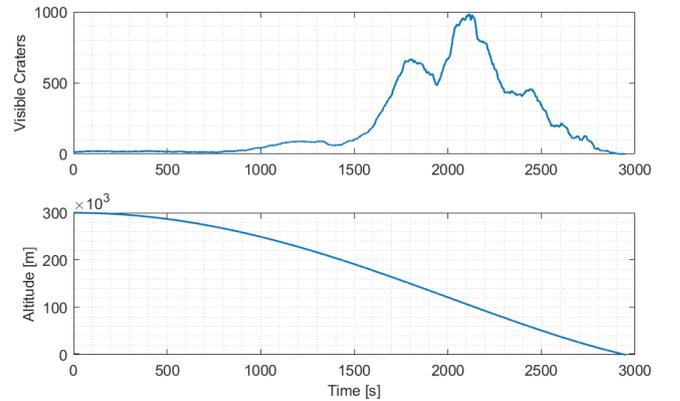

Fig. 4.   Number of visible craters to the camera and the altitude of the spacecraft.

The change in the number of visible craters can be explained easily. As the spacecraft is further away from the Moon, cameras can see a larger portion of the Moon. However, craters must be large enough to be successfully detected from high altitudes. As the spacecraft gets closer to the Moon, the visible area of the Moon to the camera decreases but craters with smaller diameters can be detected. Therefore, the number of detected craters increases. Nevertheless, when the spacecraft is too close to the surface, the number of detected craters decreases as a tiny portion of the surface is visible to the camera.

For the position and the attitude estimation, craters must be identified. Number for detected craters after the identification process and the position estimation errors are given in Fig. 5. By comparing Fig. 4. and Fig. 5., effect of the identification process can be observed. Number of craters that can be used for estimation is lower than the number of craters that is visible to the camera. Also, the position estimation accuracy of the estimation algorithm can be seen from Fig. 5.

It can be observed that the number of craters used to estimate position affects the accuracy of estimation significantly. In the first 1500 seconds, the estimation accuracy is relatively lower than the estimation accuracy between the 1500th and the 2850th seconds. After the 2850th second, the estimation accuracy declines sharply with decreasing number of craters. It can be seen that estimation is not even possible at some point.

Additionally, attitude estimation errors are plotted in Fig. 6. In order to estimate the attitude, detected craters and estimated positions given in Fig. 5 are used.

The attitude estimation results are similar to the position estimation results. As the detected crater number increases, the estimation accuracy also increases. This result is expected as the attitude estimation algorithm uses the estimated positions.

For the first 1500 seconds, the position estimation error is bounded within 30 meters in all axes, and the attitude error is bounded within 0.02 degrees for each Euler angle. Between the 1500th and the 2850th seconds, the position estimation error is bounded within 2 meters in all axes, and the attitude error is bounded within 0.002 degrees for each Euler angle.

The error increases significantly in the last 150 seconds for both position and attitude estimations. Therefore, the proposed algorithm is unsuitable to use just before the landing. In order to make a successful estimation while the spacecraft is close to the surface, inertial navigation estimation of the spacecraft must not drift much from actual states for at least 150 seconds, as the errors cannot be corrected.

Although the estimation is very accurate between the 1500th and the 2850th seconds, the algorithm uses between 500 to 800 craters for estimation. In a practical sense, image processing, crater identification and the proposed estimation algorithm cannot process information as large as this in a limited time. The maximum number of craters that can be used for estimation depends on the camera properties and processing power available on the spacecraft.

In order to observe the effects of the maximum crater limitation, simulation of the spacecraft is done by limiting the maximum number of craters used to estimate the position and the attitude. The maximum number of craters is limited to 10, 20, 50, 100 and 200 for each simulation run. Position estimation errors for each case for the first 1500 seconds are plotted in Fig. 7. for comparison.

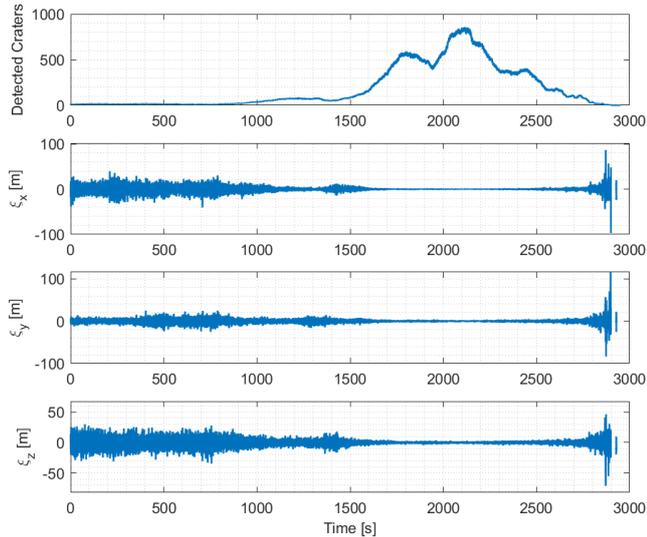

Fig. 5. Position estimation errors.

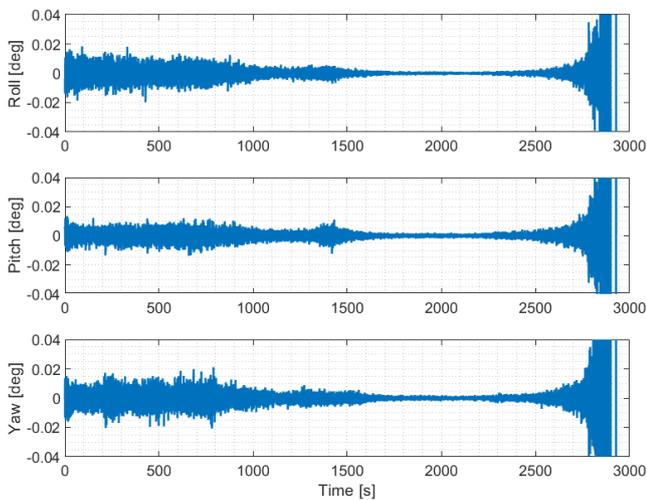

Fig. 6. Attitude estimation errors.

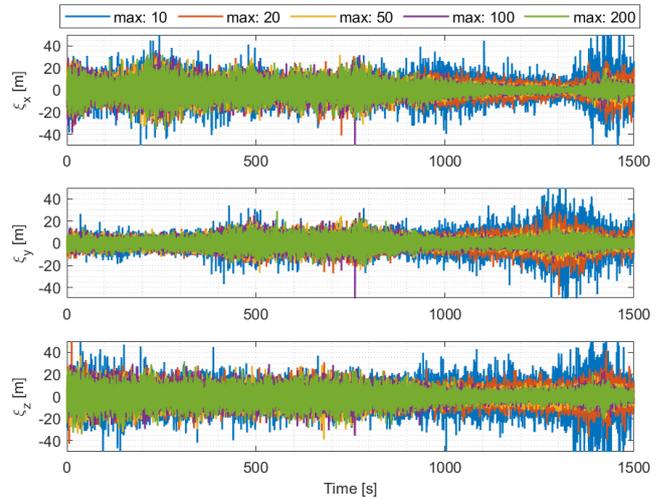

Fig. 7. Position estimation errors with limited number of craters for first 1500 seconds.

At the beginning of the simulation, limiting the number of craters that is used for estimation did not affect the estimation accuracy much. That is because the number craters visible from high altitudes is low and below the limit for most cases. Therefore, number of craters that are used for estimation are not limited for this region. However, as the spacecraft gets closer to the surface, limit shows its effect and estimation accuracy changes depending on the maximum number of craters. However, in general, this region is not the region of interest.

In order to effectively see the limitation effect, region where the visible craters are more than the maximum limit must be chosen. Therefore, using the visible crater information given in Fig. 4, estimation errors between the 1800th and the 2400th seconds are decided to be inspected and plotted in Fig. 8.

Effect of the limitation is clearly visible in Fig. 8. When the number of used craters for estimation is larger, position estimation is more accurate. However, it can also be seen that accuracy of the estimation is not increased when the maximum number of craters change from 100 to 200. Therefore, it can be concluded that using more craters than necessary will not increase the accuracy effectively. Maximum number of craters must be selected by considering how accurate the estimation is required to be.

To see the error characteristics, root mean square of the errors (RMSE) between the 1800th and the 2400th seconds are found and tabulated in Table 1.

Increasing the number of craters used for estimation improves estimation accuracy. However, the improvement is not linear. It can be seen from Table 1 that increasing the number of craters from 10 to 20 and 20 to 50 increases the estimation accuracy significantly. Nevertheless, estimation accuracy improves slightly when the number of craters increases from 50 to 100 and 100 to 200. Even if the crater number is not limited, accuracy is limited. Therefore, one must choose the maximum number of craters considering the fact that the increase in accuracy might not be as much as increase in the required computational power.

## 6. Conclusion

In this paper, an absolute pose estimation algorithm using visual information captured by the camera is designed. The algorithm uses the nonlinear least squares method for position estimation and the QUEST method for attitude estimation. In order to test the algorithm, a scenario in which the spacecraft makes a hard landing on the Moon is simulated. Using the simulation results, the accuracy of the estimation is inspected. It is seen that estimation accuracy depends on the number of craters used for estimation. However, using hundreds of crater information for estimation is not possible in a practical sense. Therefore, the maximum number of craters used for estimation is limited, and the accuracy of the estimation is inspected. It is found that increasing the maximum limit did not increase the accuracy linearly. Therefore, it is found that the maximum number of crater limit must be chosen considering that increase in accuracy might not be as much as the increase in required computational power.

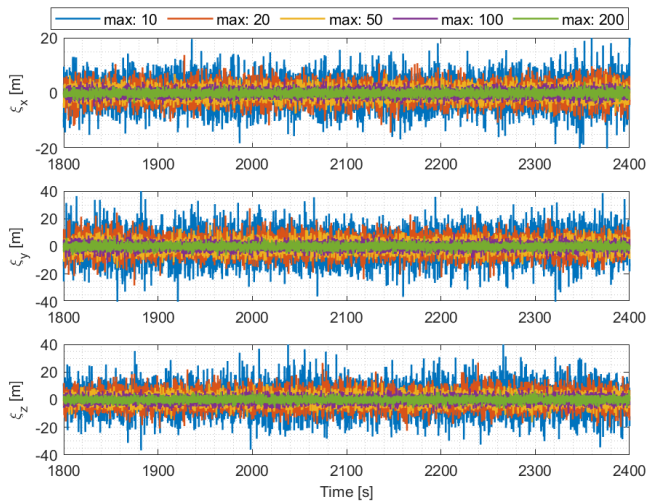

Fig. 8. Position estimation errors with limited number of craters between the 1800th and the 2400th seconds

Table 1. Root mean square of the position and attitude estimation errors

| | | Number of craters used for estimation | | | | | |
|---|---|---|---|---|---|---|---|
| | | 10 | 20 | 50 | 100 | 200 | no Limit |
| [m] | X | 4.923 | 3.2534 | 1.9981 | 1.4021 | 0.9739 | 0.6111 |
| | Y | 9.7451 | 6.2719 | 3.877 | 2.6237 | 1.8795 | 1.1499 |
| | Z | 9.3482 | 6.1251 | 3.6869 | 2.6092 | 1.8001 | 1.1181 |
| [°] | $\phi$ | 0.0039 | 0.0026 | 0.0016 | 0.0011 | 0.00079 | 0.00049 |
| | $\theta$ | 0.0066 | 0.0042 | 0.0025 | 0.0018 | 0.0012 | 0.00076 |
| | $\psi$ | 0.0066 | 0.0042 | 0.0025 | 0.0018 | 0.0012 | 0.00076 |